\documentclass[letterpaper]{article} 
\usepackage{aaai2026}  
\usepackage{times}  
\usepackage{helvet}  
\usepackage{courier}  
\usepackage[hyphens]{url}  
\usepackage{graphicx} 
\usepackage{natbib}  
\usepackage{caption} 
\usepackage{algorithm}
\usepackage[noend]{algorithmic}

\usepackage{amssymb}
\usepackage{amsmath}
\usepackage{mathtools}
\usepackage{algorithm}
\usepackage{cleveref}
\usepackage{graphicx}

\usepackage{multirow}
\usepackage{amsthm,amsmath,amsfonts}
\usepackage{caption}
\usepackage{xcolor}
\usepackage{url}
\usepackage{rotating}
\usepackage{float}
\usepackage{subfigure}
\usepackage{subcaption}
\usepackage{lineno}
\usepackage{adjustbox}
\usepackage{makecell}

\usepackage{newfloat}
\usepackage{listings}
\DeclareCaptionStyle{ruled}{labelfont=normalfont,labelsep=colon,strut=off} 
\floatstyle{ruled}
\newfloat{listing}{tb}{lst}{}
\floatname{listing}{Listing}
\title{From Stories to Cities to Games: A Qualitative Evaluation of Behaviour Planning}

\author{
    Mustafa F. Abdelwahed\textsuperscript{1},
    Joan Espasa\textsuperscript{1},
    Alice Toniolo\textsuperscript{1},
    Ian P. Gent\textsuperscript{1}
}
\affiliations{
    \textsuperscript{\rm 1}University of St Andrews, School of Computer Science, UK\\
    \{ma342, jea20, a.toniolo, ian.gent\}@st-andrews.ac.uk
}

\newtheorem{definition}{Definition}

\newcommand{\ma}[1]{\textcolor{black}{#1}}

\newcommand{\fbixoperator}[1]{\texttt{FBI}_\texttt{#1}}

\newcommand{\plangenerator}[1]{\operatorname{PlanGenerator}_{#1}}
\newcommand{\behaviourgenerator}[1]{\operatorname{BehaviourGenerator}_{#1}}

\newcommand{\planningtask}{\Xi}
\newcommand{\allstate}{S}
\newcommand{\state}[1]{s_{#1}}
\newcommand{\allaction}{A}
\newcommand{\action}[1]{a_{#1}}
\newcommand{\transitionfn}{\gamma}
\newcommand{\costfn}{\operatorname{cost}}
\newcommand{\initialstate}{I}
\newcommand{\goalstate}{G}
\newcommand{\singleplan}{\pi}
\newcommand{\allplans}{\Pi_\planningtask}
\newcommand{\diversesetplan}[1]{\Psi_{#1}}
\newcommand{\planscount}{k}

\newcommand{\basicformula}[1]{\phi_{#1}}
\newcommand{\smtmodel}[1]{\mathcal{M}_{#1}}

\newcommand{\featureexpr}[1]{\textit{Expr}_{#1}}

\newcommand{\dimensionlist}{\Delta_\planningtask}
\newcommand{\dimension}[1]{\Delta_{#1}}

\newcommand{\extractfn}[1]{\odot_{#1}}

\newcommand{\diversityfeatureslist}{F_{\planningtask}}
\newcommand{\diversityfeature}[1]{f_{#1}}

\newcommand{\planbehaviour}{\operatorname{PBehaviour}}
\newcommand{\behaviourcount}{\operatorname{BDC}}

\begin{document}

\maketitle

\begin{abstract}

The primary objective of a diverse planning approach is to generate a set of plans that are distinct from one another. Such an approach is applied in a variety of real-world domains, including risk management, automated stream data analysis, and malware detection. More recently, a novel diverse planning paradigm, referred to as behaviour planning, has been proposed. This approach extends earlier methods by explicitly incorporating a diversity model into the planning process and supporting multiple planning categories. In this paper, we demonstrate the usefulness of behaviour planning in real-world settings by presenting three case studies. The first case study focuses on storytelling, the second addresses urban planning, and the third examines game evaluation.
\end{abstract}



\section{Introduction}

Classical planners focus on generating a single solution, referred to as a plan. In contrast, diverse planners produce multiple plans that are distinct from one another for a given task. There are several reasons why a user may require multiple plans. While \citet{haessler1991cutting} showed that generating several different solutions can help account for future situations, \citet{nguyen2012generating} demonstrated that generating different plans is beneficial because such plans may account for side information (e.g., user preferences) that hard to model in the problem. In practice, the generated solution can sometimes be challenging to implement; thus, offering a set of alternative solutions is often more useful, enabling users to choose among different viable options~\cite{ingmar2020modelling,cully2017quality}.
From an application viewpoint, various real-world applications use diverse planning approaches as a core component. For instance, \citet{sohrabi2018ai} applied diverse planning to anticipate a range of potential future scenarios from a risk management perspective. In this setting, diverse planners are used to generate and rank multiple expected future scenarios. Another important application domain is malware detection~\cite{boddy2005course,sohrabi2013hypothesis}, where diverse planners are employed to identify malicious activity within network streams. In this context, a planner attempts to generate plans that explain observed network behaviour in order to detect attacks. Using diverse planning techniques enables the detection of multiple potential malicious activities rather than focusing on a single explanation. Beyond these domains, diverse planning also underpins several other applications, including pipeline generation and learning in machine learning~\cite{katz2020exploring} and business process automation~\cite{chakraborti2020robotic}, further highlighting its broad applicability.

Several diverse planning frameworks have been proposed to generate $\planscount$ optimal or sub-optimal diverse plans~\cite{srivastava2007domain,roberts2013tale,vadlamudi2016combinatorial}. The most recent approach is proposed by \citet{abdelwahed2024behaviour} and is referred to as \emph{behaviour planning}. In this approach, diversity is represented using \emph{behaviour spaces}, a concept inherited from the Quality-Diversity Optimisation field~\cite{lehman2011abandoning}.
A behaviour space is an n-dimensional grid in which each dimension corresponds to a feature of interest to the user, and each cell within the grid is referred to as a \emph{behaviour}. The diversity planning problem is formulated as the task of finding a set of plans that maximises the number of distinct behaviours. Behaviour planning comprises two primary components: the Behaviour Sorts Suite (\texttt{BSS}) and Forbid Behaviour Iterative ($\fbixoperator{}$). The former is a qualitative framework used to describe the diversity model, while $\fbixoperator{}$ is a planning approach that uses the diversity model defined by \texttt{BSS} to generate diverse plans. Two implementations of behaviour planning are available. The first~\cite{abdelwahed2024behaviour} is based on planning-as-satisfiability and targets planning problems modelled using declarative languages such as PDDL (model-based); this implementation is referred to as $\fbixoperator{SMT}$. The second~\cite{abdelwahed2025diverse} targets planning problems that are modelled using simulators and does not require an explicit domain model (model-free); this implementation is referred to as $\fbixoperator{LTL}$. 

In this paper, we present three real-world case studies that demonstrate the benefits of behaviour planning. These cases are implemented\footnote{The code for the case studies can be found at \url{https://github.com/MFaisalZaki/behaviour-planning-case-studies.git}.} and their output is presented in this paper. The first case study explores the use of behaviour planning for generating diverse narratives in storytelling~\cite{rivera2024story}. In this setting, we employ $\fbixoperator{SMT}$ to generate a set of distinct story plots, where diversity is defined in terms of the values of a preselected set of fluents in the goal state.
The second case study focuses on urban planning and investigates the generation of diverse plans for this domain~\cite{su12030797}. Here, we apply $\fbixoperator{LTL}$ to produce alternative layouts for the Town of St Andrews, with plan diversity characterised by factors such as sustainability and diversity.
The final case study delves into the concept of game replayability~\cite{game-diversity-measure}. In this case study, we employ $\fbixoperator{LTL}$ to assess the replayability of the classic Super Mario Land (Game Boy version) by generating a variety of actions that lead to winning the 1-1 world. We distinguish between two types of plans for this case study: those that involve Super Mario killing or avoiding an enemy (e.g., Gomba). These benefits are expressed from a user's viewpoint. \citet{abdelwahed2024behaviour} employed three personas proposed by~\citet{sreedharan2020emerging} to represent users for behavior planning. These personas were: (i) the end user, who interacts with the planning system through a user interface; (ii) the domain expert, who sets high-level mission objectives for the planning system; and (iii) the algorithm designer, who generates plans based on the end user and domain designer's requirements. In this work, we will cover each case study's end user separately. Regarding the domain expert, we consider ourselves as the domain expert, and the algorithm designer will be \citet{abdelwahed2024behaviour} and \citet{abdelwahed2025diverse} since we are using their implementations.
The remainder of the paper is organised as follows. We first provide background on the diversity planning problem and behaviour planning. We then present the three case studies in detail. Finally, we conclude with a discussion of limitations, possible improvements, and directions for future work.

\begin{figure*}[]
    \centering
    \includegraphics[scale=0.65]{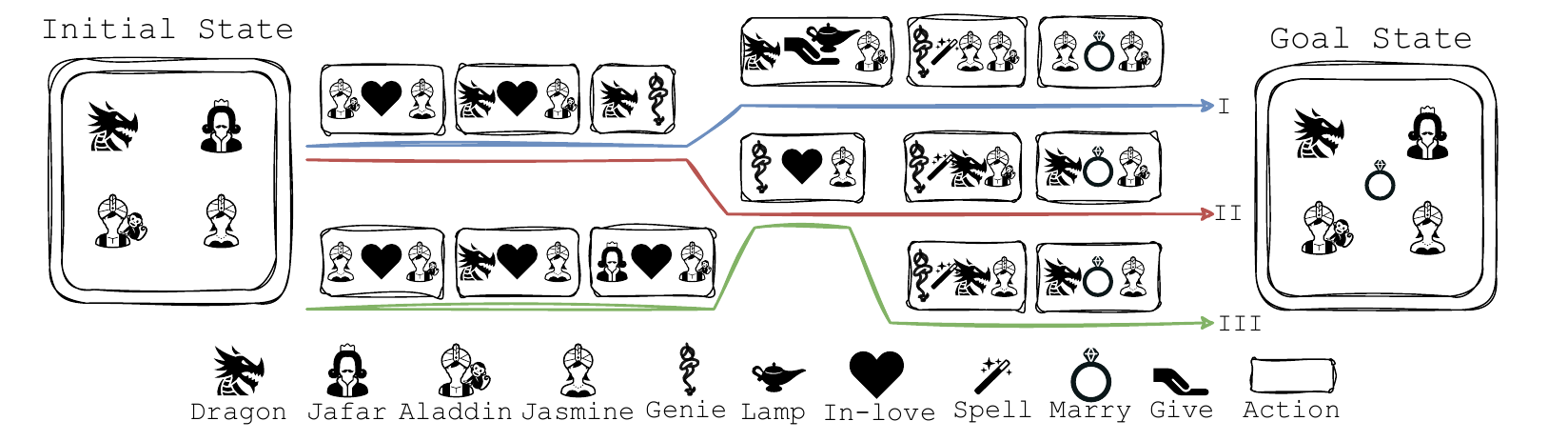}
    \caption{Three diverse narratives for the Aladdin world.}
    \label{fig:aladdin-narratives}
\end{figure*}

\section{Background}
This section provides the necessary background for our approach. We first define the classical planning problem, then introduce the diversity model used to represent and quantify plan diversity, and finally describe the behaviour planning framework that generates diverse plan sets.

\subsection{Planning Problem}
Inspired by \citet{ghallab2016automated}, we define the planning problem as follows.

\begin{definition}[Planning Problem]\label{def:planning-problem}
    A planning problem is a tuple $\planningtask=\langle \allstate, \allaction, \transitionfn, \costfn, \initialstate, \goalstate\rangle$, where:
    \begin{itemize}
        \item $\allstate$ is a finite set of states
        \item $\allaction$ is a finite set of actions
        \item $\transitionfn: \allstate \times \allaction \rightarrow \allstate$ is a transition function that maps each state $\state{}\in \allstate$ and action $\action{}\in \allaction$ to a successor state $\transitionfn(\state{},\action{})=\state{}^\prime$
        \item $\costfn: \allaction \rightarrow \mathbb{R}^{+}$ is a cost function assigning a non-negative cost to each action
        \item $\initialstate\in \allstate$ is the initial state, and
        \item $\goalstate$ is a formula representing the goal condition.
    \end{itemize}
\end{definition}

A \emph{plan} $\singleplan$ is a sequence of actions $\singleplan = \langle\action{1}, \action{2}, \ldots, \action{m}\rangle$ such that each $\action{i}\in \allaction$ and executing $\singleplan$ from $\initialstate$ yields a state satisfying $\goalstate$; that is, $\transitionfn(\ldots\transitionfn(\transitionfn(\initialstate, \action{1}),\action{2})\ldots,\action{m}) \models \goalstate$. We denote by $\allplans$ the set of all valid plans for a given planning problem $\planningtask$. The cost of a plan is defined as the sum of its action costs: $\costfn(\singleplan) = \sum_{i=1}^{m} \costfn(\action{i})$.
%
While our formulation includes the cost function for generality, we note that in this work we focus on plan diversity rather than optimality, and therefore do not enforce cost minimisation.

\subsection{Diversity Model and Behaviour Count}
\begin{figure}[H]
    \centering
    \includegraphics[scale=1.1]{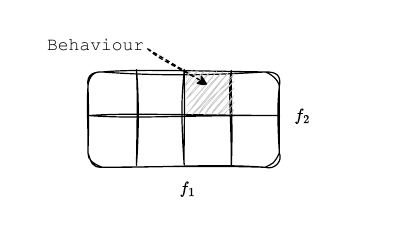}
    \caption{A 2-dimension behaviour space illustration. }
    \label{fig:behaviour-space-example}
\end{figure}

To represent and measure diversity among plans, we employ a diversity modelling approach called the \emph{Behaviour Sorts Suite} (\texttt{BSS}), proposed by \citet{abdelwahed2024behaviour}. This model represents diversity using an $n$-dimensional grid, called the \emph{behaviour space}, where each dimension corresponds to a feature of interest. Each point (or cell) within this grid corresponds to a unique \emph{behaviour}. \Cref{fig:behaviour-space-example} shows an illustration of a behaviour space with two features $\diversityfeature{1}$ and $\diversityfeature{2}$, and the shaded box denotes a behaviour. The diversity of a set of plans is quantified by counting the number of distinct behaviours represented in the set, via a metric called the \emph{behaviour diversity count}.

Formally, let $F_\planningtask=\{\diversityfeature{1},\ldots,\diversityfeature{n}\}$ be a set of user-defined features. Each feature $\diversityfeature{i}=\langle \dimension{i}, \extractfn{i}, \featureexpr{i}\rangle$ consists of:
\begin{itemize}
    \item a domain $\dimension{i}$ containing the possible values for feature $\diversityfeature{i}$
    \item an extraction function $\extractfn{i}:\allplans\rightarrow \dimension{i}$ that computes the value of feature $\diversityfeature{i}$ for a given plan, and
    \item a feature expression $\featureexpr{i}$, a first-order logic formula with equality that specifies how to compute the feature value from the plan
\end{itemize}

The behaviour space is the Cartesian product of all feature domains: $BS_{\dimensionlist}= \dimension{1} \times \dimension{2} \times \ldots \times \dimension{n}$. We collect all extraction functions into a set $\extractfn{\dimension{}} = \{\extractfn{1}, \ldots, \extractfn{n}\}$.

A \emph{plan behaviour} is an $n$-tuple$\langle\delta^1,\ldots,\delta^n\rangle$ where each $\delta^i\in\dimension{i}$ represents the value of feature $\diversityfeature{i}$. Given a plan $\singleplan$ and the extraction functions $\extractfn{\dimension{}}$, the behaviour of $\singleplan$ is computed as:
\[
\planbehaviour(\extractfn{\dimension{}}, \singleplan)=\langle \extractfn{1}(\singleplan), \extractfn{2}(\singleplan),\ldots,\extractfn{n}(\singleplan)\rangle.
\]

The \emph{behaviour diversity count} of a set of plans $\diversesetplan{\planningtask} \subseteq \allplans$ is the number of distinct behaviours in the set:
\[
\behaviourcount(\extractfn{\dimension{}},\diversesetplan{\planningtask})=\left| \{ \planbehaviour(\extractfn{\dimension{}}, \singleplan) \mid \singleplan \in \diversesetplan{\planningtask} \} \right|.
\]

For further details on the Behaviour Sorts Suite, we refer the reader to \citet{abdelwahed2024behaviour}.

\subsection{Diversity Planning Problem}
Building upon the planning problem (\Cref{def:planning-problem}), we now define the diversity planning problem, which seeks a set of plans rather than a single plan. Our formulation is a modification of the diversity planning problem defined by \citet{abdelwahed2024behaviour}, adapted to focus on behavioural diversity without enforcing cost optimality.

\begin{definition}[Diversity Planning Problem]\label{def:diversity-planning-problem}
    Given a planning problem $\planningtask$, a required number of plans $\planscount$, and a set of feature dimensions $\dimensionlist$, find a set of plans $\diversesetplan{\planningtask}\subseteq\allplans$ such that:
    \begin{enumerate}
        \item $\vert\diversesetplan{\planningtask}\vert\leq \planscount$, and
        \item the behaviour diversity count $\behaviourcount(\extractfn{\dimension{}},\diversesetplan{\planningtask})$ is maximised.
    \end{enumerate}
\end{definition}

\subsection{Behaviour Planning}
Following \citet{abdelwahed2024behaviour}, we use a planning approach called \emph{Forbid Behaviour Iterative} ($\texttt{FBI}_\texttt{X}$), which generates diverse plans based on the \texttt{BSS} diversity model. The subscript $\texttt{X}$ denotes the specific implementation used for the underlying solver.

Given a behaviour space $BS_{\dimensionlist}$ for a planning task $\planningtask$, $\texttt{FBI}_\texttt{X}$ iteratively generates plans with novel behaviours: it finds a plan, forbids its behaviour from future iterations, and repeats until the required number of plans is obtained. \Cref{alg:fbi-planner-main} outlines this procedure.

The algorithm begins with an empty plan set $\diversesetplan{\planningtask}$ (Line~\ref{alg-line:bspace-planner-initial-set}). The first loop (Lines~\ref{alg-line:fbi-main-loop-start}--\ref{alg-line:fbi-main-loop-end}) iteratively generates plans with previously unseen behaviours using the $\operatorname{BehaviourGenerator}_\texttt{X}$ function, continuing until either no new behaviours can be found or $\planscount$ plans have been collected. If the required number of plans $\planscount$ exceeds the number of available behaviours (i.e., $\planscount > \vert BS_{\dimensionlist}\vert$), a second loop (Lines~\ref{alg:fbi-k-start-loop}--\ref{alg:fbi-k-end-loop}) generates additional plans using $\operatorname{PlanGenerator}_\texttt{X}$, which produces plans without the behaviour novelty constraint.

\begin{algorithm}
\caption{$\texttt{FBI}_\texttt{X}$~\cite{abdelwahed2024behaviour}}\label{alg:fbi-planner-main}
\begin{algorithmic}[1]
\REQUIRE $\planningtask$: Planning task, $\diversityfeatureslist$: Diversity features, $\planscount$: Required number of plans
\ENSURE $\diversesetplan{\planningtask}$: Set of plans with different behaviours, $\behaviourcount$: Behaviour diversity count
\STATE $\diversesetplan{\planningtask} \gets \emptyset$; $\behaviourcount\gets 0$\label{alg-line:bspace-planner-initial-set}

\WHILE{$\vert\diversesetplan{\planningtask}\vert <\planscount$} \label{alg-line:fbi-main-loop-start}
\STATE $\singleplan \gets \behaviourgenerator{X}(\planningtask, \diversityfeatureslist, \diversesetplan{\planningtask})$
\IF{$\singleplan \neq \varnothing$}
\STATE $\diversesetplan{\planningtask} \gets \diversesetplan{\planningtask} \cup \{\singleplan\}$
\STATE $\behaviourcount \gets \behaviourcount+1$
\ELSE
\STATE \textbf{break}
\ENDIF
\ENDWHILE \label{alg-line:fbi-main-loop-end}

\WHILE{$\vert\diversesetplan{\planningtask}\vert <\planscount$} \label{alg:fbi-k-start-loop}
\STATE $\singleplan\gets \plangenerator{X}(\planningtask,\diversesetplan{\planningtask})$
\IF{$\singleplan = \varnothing$}
\STATE \textbf{break}
\ENDIF
\STATE $\diversesetplan{\planningtask} \gets \diversesetplan{\planningtask} \cup \{\singleplan\}$
\ENDWHILE \label{alg:fbi-k-end-loop}
\RETURN $\diversesetplan{\planningtask}, \behaviourcount$
\end{algorithmic}
\end{algorithm}

\Cref{alg:fbi-planner-main} is a general framework that can accommodate different implementations of the behaviour space and plan generation. In this work, we use two implementations from the literature.

The first implementation, proposed by \citet{abdelwahed2024behaviour}, uses \emph{planning-as-satisfiability} to represent the behaviour space and generate plans. In this approach, the planning problem is encoded as a propositional satisfiability formula $\basicformula{n}$, where $n$ denotes the plan horizon. A satisfying assignment $\smtmodel{\basicformula{n}}$ represents a valid plan, and we use the notation $\smtmodel{\basicformula{n}}[v]$ to extract the value of a variable $v$ from the assignment.
The second implementation, proposed by \citet{abdelwahed2025diverse}, uses \emph{Linear Temporal Logic} (LTL)~\cite{rozier2011linear} to specify the behaviour space. Each dimension is expressed as an LTL formula, and a behaviour corresponds to the conjunction of these dimension-specific formulae.
The specific dimension implementations used in this work are described in the following section.


\section{Case Studies}
This section covers each case study regarding its background, the corresponding behaviour space dimensions, and the resulting generated plans. 


\subsection{Storytelling}
This case study focuses on narrative generation, which is commonly described as comprising three layers: plot, discourse, and narration~\cite{rivera2024story}. In this work, we concentrate on plot planning. A plot planner constructs a timeline of character actions that transform the virtual world from its initial configuration to the author’s desired goal, referred to as the plot. A diverse planner can generate multiple plots for a given story, enabling the author to select the one that best aligns with their preferences.
One application of diverse planning in this context is the generation of multiple plots that lead to a specific ending. Alternatively, diverse planning can be used to generate different possible endings. In this case study, we focus on the latter. To the best of our knowledge, no existing diverse planner has been specifically designed to support narrative generation. \citet{riedl2010narrative} proposed the use of planning in story generation to ensure that characters exhibit intentional behaviour, where actions are driven by explicit motivations. These challenges are typically addressed using a specialised class of planners known as intentional planners. To overcome this issue we used the compilation suggested by~\citet{haslum2012narrative} to convert intentional planning problem into classical planning one.

\begin{figure}[htbp]
    \centering
    \includegraphics[scale=1.1]{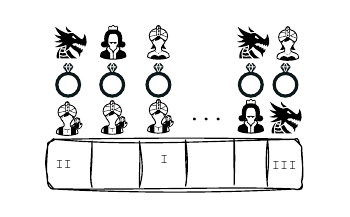}
    \caption{Behaviour space for the Aladdin story domain, illustrating how different narrative outcomes map to distinct behaviours.}
    \label{fig:aladdin-bs}
\end{figure}

For this case study we will use the Aladdin story domain suggested by~\citet{riedl2010narrative}. This narrative features five characters: Aladdin, Jasmine, Genie, Jafar and Dragon. Any character can move between locations, trade the genie lamp if they have it, cast love spells if they control the genie, and marry another character if they are in love. In this domain, we formulate the problem as finding a set of narratives such that the difference between them is based on whom gets married at the end of the story. Therefore, we define the goal state as a disjunction of the marry fluent (i.e., \texttt{ (exists (?c1 - char ?c2 - char) (married-to ?c2 ?c1))}). The feature for this case study is called possible-endings ($\diversityfeature{pe}$). Its corresponding dimension includes $\dimension{pe}=\{g|g\subseteq\operatorname{gnd}(G)\}$\footnote{We use the $\operatorname{gnd}$ operator to ground a given formula.} a set of all possible grounded predicates in the goal state and $\extractfn{pe}$ which returns the values of grounded fluents at the goal state. For the Aladdin story, $\dimension{pe}$ will be $\{\texttt{married-to(Aladdin, Jasmine)}, \allowbreak \ldots, \allowbreak \texttt{married-to(Dragon, Jafar)}\}$. The extracting function $\extractfn{pe}$ returns the values of the fluents in $\dimension{pe}$ at the goal state. The $\featureexpr{pe}$ will be the conjunction of the assignments of the fluents in $\operatorname{gnd}(G)$ (i.e., $\featureexpr{pe} = \bigwedge_{g\in\operatorname{gnd}(G)} \left(g=\smtmodel{\basicformula{n}^\prime}[g]\right)$).

\Cref{fig:aladdin-narratives} shows three narratives generated by $\fbixoperator{SMT}$ based on the specified dimension. The first narrative depicts Aladdin falling in love with Jasmine; the dragon then falls in love with Aladdin, which leads it to give them the lamp. Aladdin subsequently uses the lamp to cast a love spell on Jasmine in order to marry them. The other two narratives depict scenarios in which all characters fall in love with one another, and the genie ultimately decides who marries the dragon. \Cref{fig:aladdin-bs} maps these generated narratives into a behaviour space.

In this context, the end user would be the story author, who benefits from behaviour planning by exploring diverse narrative outcomes without the need for manual creation. behaviour planning generates plans that encompass these possibilities based on predefined dimensions. As for the domain expert, we introduced a new dimension based on the narrative designer’s model. It’s important to note that the generated narratives do not account for character believability, which is a consequence of the compilation used in this case study. Incorporating character believability would require a different compilation approach, but this is beyond the scope of this paper.

\subsection{Urban Planning}

\begin{figure}
    \includegraphics[scale=0.55]{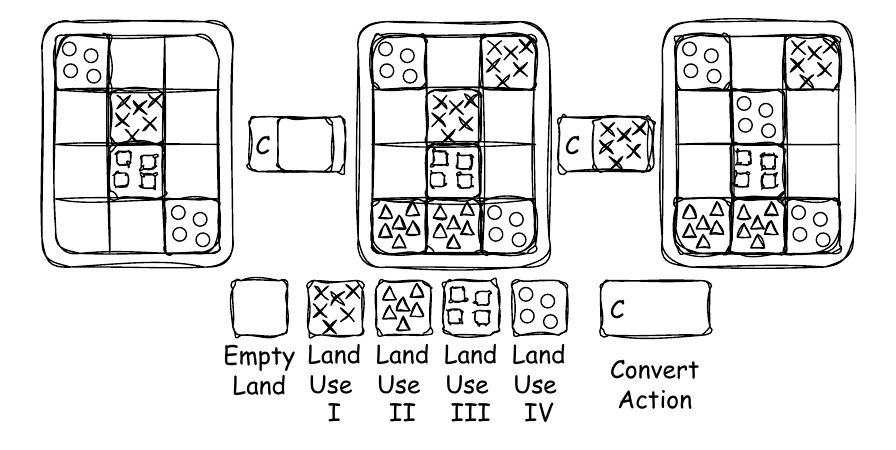}
    \caption{A simple illustration of the urban planning simulator operation.}\label{fig:urban-planning-simulator}
\end{figure}

\begin{figure}[!b]  
\centering
\subfigure[Current land use.]{\includegraphics[scale=0.22,height=0.15\paperheight]{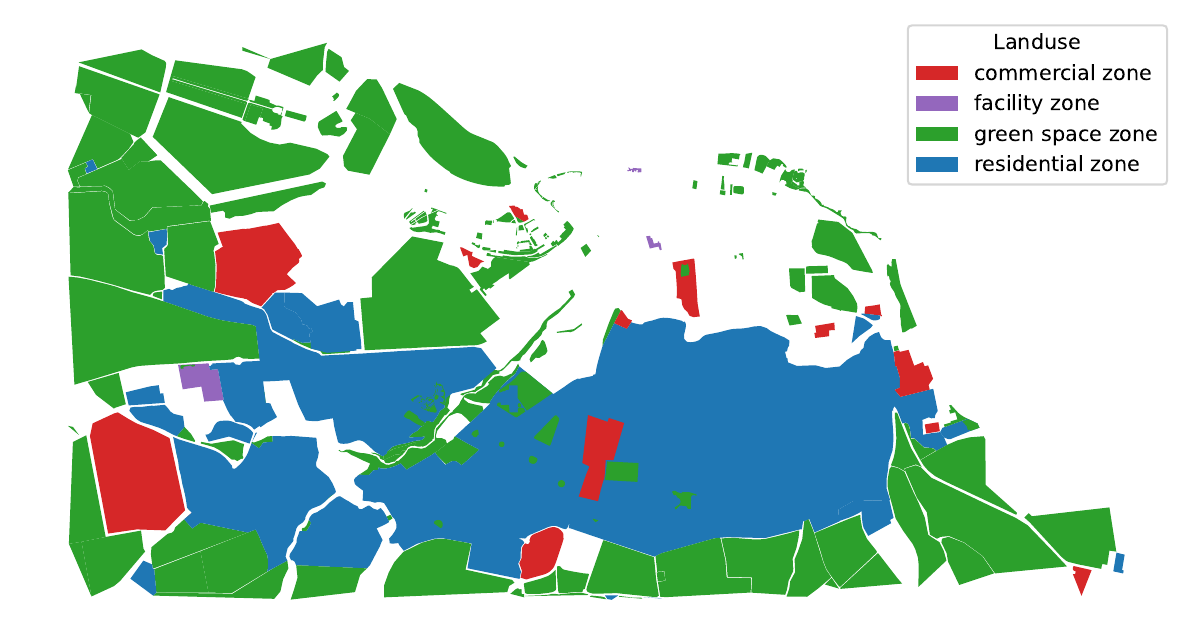}}
\subfigure[\texttt{Plan-I}: high sustainability and moderately high diversity.]{\includegraphics[scale=0.22,height=0.15\paperheight]{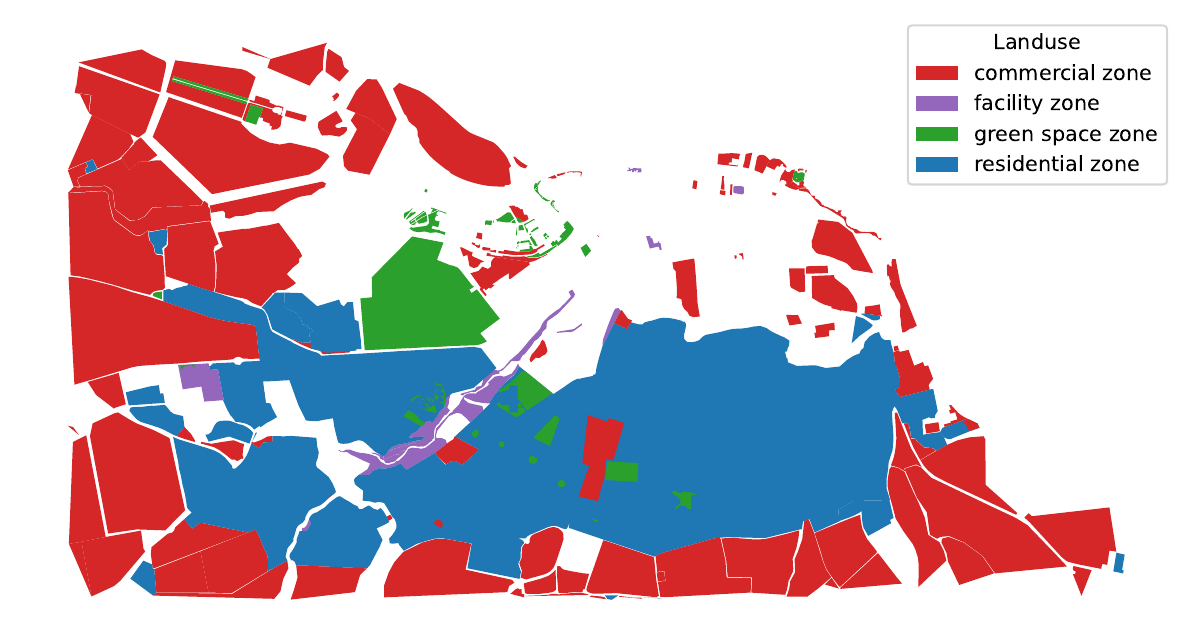}}
\subfigure[\texttt{Plan-II}: high sustainability and below average diversity.]{\includegraphics[scale=0.22,height=0.15\paperheight]{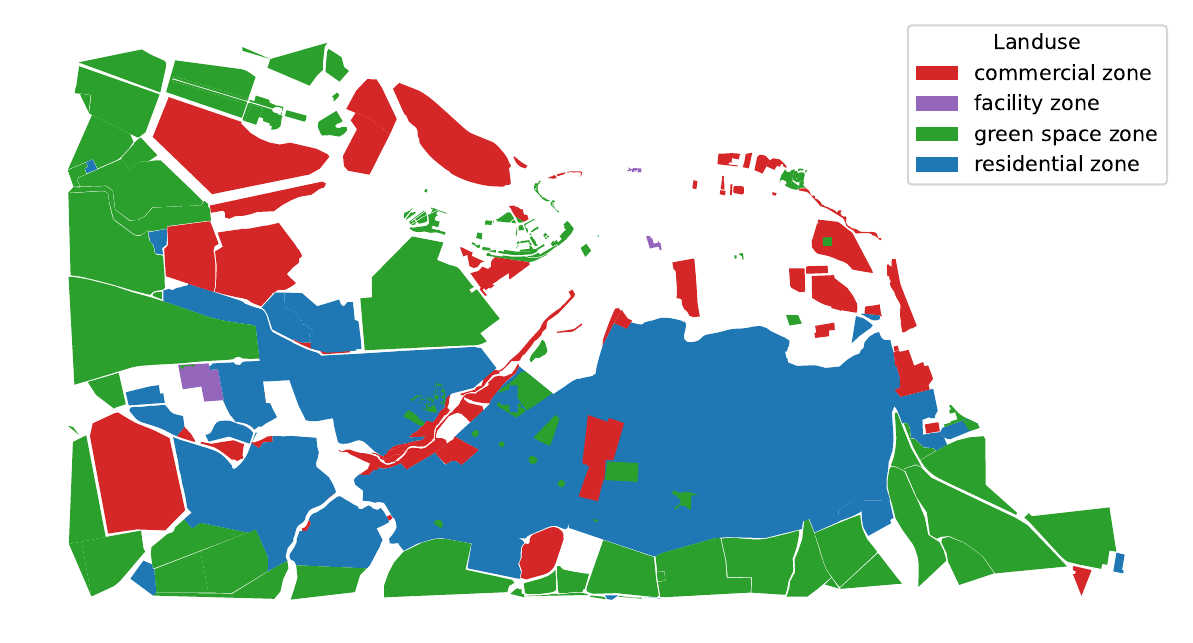}}
\caption{Urban plans for the town of St Andrews. Red color means \texttt{commercial zone}, purple means \texttt{facility zone}, green means
\texttt{green space zone}, and blue means
\texttt{residential zone}}\label{fig:urban-plans-st-andrews}
\end{figure}

Urban planning is about modifying land usage to meet residents' demands and promote sustainable~\cite{su12030797}. An urban planning expert (end user) computes the \emph{Sustainability} and \emph{Diversity} scores for the current land usage and suggest changes to improve these scores. Such scores are used to quantify the quality of an urban plan~\cite{ma-urban-planning}.
%
Sustainability seeks a balance between environmental concerns and economic growth. Its score is measured as the ratio of the green space, commercial and facilities zone to the total number of land. Regarding diversity, it aims to balance between the urban ecosystem (i.e., social, economic, and spatial characteristics). Its score is computed using the Shannon-Weaver formula. \ma{The objective for this case study is to generate different land usages that have different sustainability and diversity scores. Since there is no specific goal formula, we define the planning problem as horizon planning problem~\cite{horizon-planning-problem}. A horizon planning problem shares everything with a planning problem except for the goal formula, it is replaced with a budget (i.e., maximum plan length). In this work, we model the problem using a simulator instead of a declarative model. This decision is made due to the limitation of modelling complex operations when using a declarative approach. To clarify this, } 
%
%
%
\Cref{fig:urban-planning-simulator} illustrates the operation of the simulator. The first action asks the simulator to change empty lands on this state. Based on the conversion rules encoded into the simulator, it converted three empty lands into one \texttt{land-use-I} and two \texttt{land-use-II}. To have a more realistic impact, a proper conversion rules must be implemented in the simulator which will be based on an urban planner expert. This is a simple demonstration of urban planning using simulators. Note that this problem could be solved using Constraint Programming (CP) techniques. However, we chose to approach this problem as planning problem is that the planner can provide different transformation plans (i.e., the order of applying changes to the land usages) for the urban then the end user can pick which execution plan to follow based on their preference, while CP techniques will provide the final land usage only.

\begin{figure}
    \centering
    \includegraphics[scale=0.5]{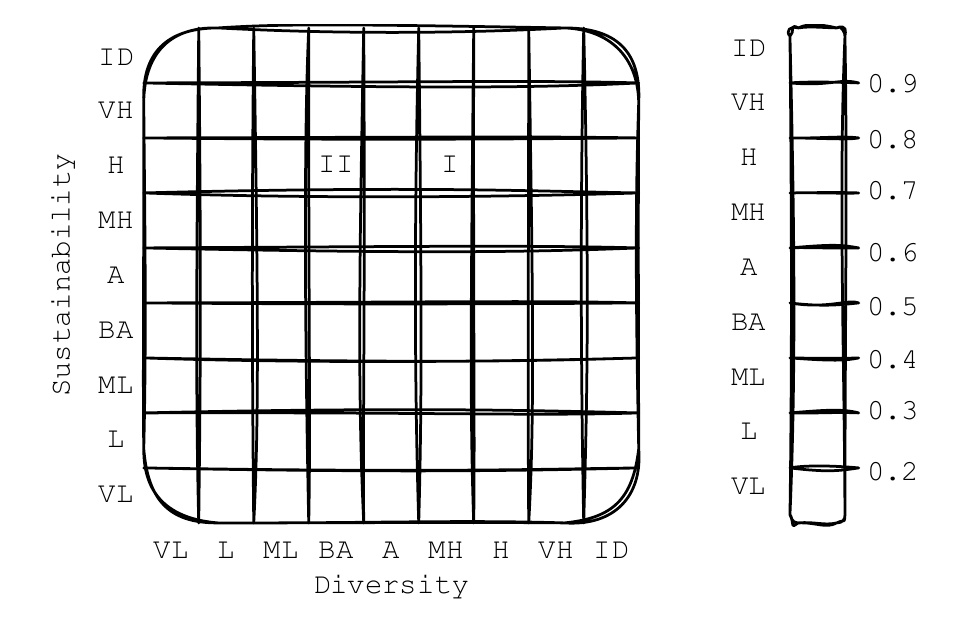}
    \caption{Behaviour space used to generate the urban plans presented in \Cref{fig:urban-plans-st-andrews}. Behaviour space is the gird represented on the left, while the categorical range is shown on the right. }
    \label{fig:urban-plan-behaviour-space}
\end{figure}

\begin{figure*}[thbp]
    \includegraphics[scale=1.35]{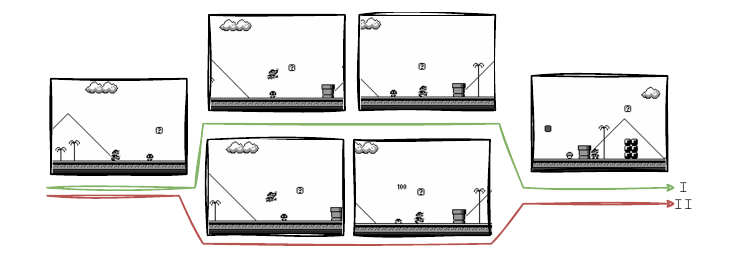}
    \caption{Two diverse plans for Super Mario world 1-1.}
    \label{fig:supermario-plans}
\end{figure*}

Following the state representation proposed by \citet{ma-urban-planning}, we formulate the problem as a diversity planning problem as finding a set of urban plans were the difference between these plans is based on their sustainability and diversity scores and each plan has $l\in\mathbb{N}^+$ actions. In this formulation, we identified five general land use types: \texttt{residential zones}, \texttt{office spaces}, \texttt{green spaces}, \texttt{commercial zones}, and {facilities}. The actions for this planning problems are converting a land type from one type to another. \ma{The simulator selects which set of lands and convert it based on the encoded rules.} We assumed a simple transformation rule for each land type. For example, if a planner would replace a green space land, then 5\% of the available green space zone will be converted into a commercial zone and facilities equally. Encoding such rules using declarative languages would be challenging, thus we are using a simulator instead. Note that the simulator can be extended further to hand pick which lands in a given type can be transformed, currently we use the top n lands in the list for each type. In this setup, we differentiate between plans based on two features: $\diversityfeature{S}$ and $\diversityfeature{D}$. The first one represents sustainability, while the second dimension represents diversity. \ma{Since $\fbixoperator{LTL}$ is based on Linear Temporal Logic, which limits it to boolean predicates. Converted the sustainability and diversity scores into categorical values and each value is represented with a boolean predicate. The sustainability and diversity scores ranges from 0 to 100, thus we define the ranges as follows: \texttt{VL} denotes that the score is $\leq 20\%$, \texttt{L} denotes that the score is between 20 and 30. \texttt{ID} represents that the score is $> 90\%$}. The corresponding dimensions for these features will be the same $\dimension{S}=\dimension{D}=\{\texttt{VL},\ldots,\texttt{ID},\texttt{l-reached}\}$. The \texttt{l-reached} is needed to notify the planner that it reached the horizon. \ma{The categorical range for $\dimension{S/D}$ is represented in \Cref{fig:urban-plan-behaviour-space} }
As for the extracting functions $\extractfn{S}$ and $\extractfn{D}$, they would compute the scores for their perspective features. Note that since there are no goal formula for this problem, we give the planner budget of 10 actions to generate an urban plan. The expression for each dimension will be $\lozenge\square\texttt{VVV}_\texttt{S/D}$, were $\texttt{VVV}\in\dimension{S/D}$. We used $\fbixoperator{LTL}$ to generate two diverse urban plans for the town of St Andrews, UK. 

\Cref{fig:urban-plans-st-andrews} illustrates the current state of the town and the two proposed plans. One plan proposes transforming some of the green spaces into commercial and facility areas to enhance diversity. In contrast, the other plan suggests altering most of the green spaces to achieve higher diversity. The end user for this case study would be an urban planner, who would benefit from behaviour planning through suggesting various urban plans with different sustainability and diversity scores. 

%

\subsection{Game Evaluation}


\citet{game-diversity-measure} showed that the diversity of game content has a significant impact on player engagement and satisfaction. This diversity encompasses procedural content generation~\cite{guzdial2022procedural}, which automates the creation of engaging content during gameplay. Current research recognises diversity as an important quality factor when assessing generated content~\cite{pcg-diversity}.
\citeauthor{game-diversity-measure} proposed considering the player's interaction with the game through capturing how players navigate and engage with the game content. This interaction is called a \emph{trace}, defined as a sequence of actions performed by the player to complete a level. In this work, we use $\fbixoperator{LTL}$ to identify diverse traces for a given level. The number of distinct traces that can be generated for a level will be a measure of the game's replayability.

To illustrate the practicality of behaviour planning in game evaluation, we selected a portion of World 1-1 from Super Mario Land as an illustrative domain. We differentiate between plans based on a single feature that indicates whether Super Mario encountered all enemies in the level or avoided them. This feature is called Enemy Engagement ($\diversityfeature{ee}$) and includes $\dimension{ee}=\{\texttt{killed},\texttt{avoided}\}$. The \texttt{killed} means that Mario has killed at least one enemy while \texttt{avoided} denotes that Mario has avoided all enemies. The extract function $\extractfn{ee}$ returns the values of each variable in $\dimension{ee}$ at the goal state. The $\featureexpr{ee}$ formula will be $\square\texttt{avoided}$ or $\lozenge\square\texttt{killed}$. The former makes sure that Mario did not kill any enemies by stating that the \texttt{avoid} variable will always be true. On the contrary, the $\lozenge\square\texttt{killed}$ means that at some point in the future Mario will kill an enemy. 

\begin{figure}
    \centering
    \includegraphics[scale=1.1]{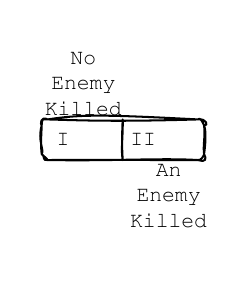}
    \caption{Behaviour space used to generate the plans presented in \Cref{fig:supermario-plans}.}
    \label{fig:super-mario-behaviour-space}
\end{figure}

To examine this, we formulated the planning problem as finding a sequence of actions such that Mario reaches the far right of the screen. Since such problem is hard to represent using a declarative language, we used an emulator called PyBoy~\footnote{\url{https://github.com/Baekalfen/PyBoy}} to emulate the game using a legally owned ROM by one of the paper's authors. \cite{abdelwahed2025diverse} highlighted that $\fbixoperator{LTL}$ can work with any tree search-based planner as long as the behaviour pruning phase is injected in the search process. Since $\fbixoperator{LTL}$ used tree search planner did not perform well, we replaced it with a specialised $\texttt{A}^\ast$ planner aimed for Super Mario. The primary reason for replacing $\fbixoperator{LTL}$'s tree search planner with an $\texttt{A}^\ast$ is that it is the best agent winning the Super Mario competition~\cite{super-mario-ai}. \Cref{fig:supermario-plans} shows that $\fbixoperator{LTL}$ generates two plans: one plan kills the Goomba (i.e., enemy), and the second one avoids it. \ma{\Cref{fig:super-mario-behaviour-space} shows the mapping of these plans to the behaviour space used to generate them.}

In this context, the game designer (end user) indirectly benefits from behaviour planning through enhanced replayability. A level with multiple, diverse solutions offers a richer gameplay experience, allowing players to experiment with strategies and achieve varying outcomes. For instance, this simple case study demonstrates that there are multiple ways to play Super Mario Land world 1-1. Additionally, other dimensions can be included, such as the game final score, the number of mushrooms Mario ate, or the number of coins collected. However, the challenge we faced was encoding such dimensions because the game version is not fully reversed, making it difficult to infer those information from the game. To win the whole level and then the game, for every portion of the map we need to plan to reach the far left of the screen and then execute the plan and keep this loop until Mario reaches the end of this level. 
\section{Conclusions \& Future work}

In this paper, we investigated the applicability of behaviour planning as diverse planning framework across several real-world case studies. These case studies were storytelling, urban planning, and game evaluation. For each case study, we presented an example for a problem formulation, diversity model and its implementation, and which behaviour planning implementation to use it with. 
The storytelling case study illustrated how behaviour planning can be leveraged to generate multiple narrative outcomes from the same initial conditions, offering authors structured control over story diversity without requiring manual plot engineering. The urban planning case study demonstrated that behaviour planning with simulators can effectively explore alternative urban development plans under while considering factors such as sustainability and diversity. Finally, the game evaluation case study showed how behaviour planning can be used as an analytical tool to assess replayability by identifying distinct player traces, providing a behaviour-level perspective on content diversity rather than focusing solely on structural game features.

These case studies show that behaviour planning is domain-agnostic and supports various planning categories (i.e., model-based/free). Rather than treating diversity as a secondary optimisation criterion or a post-processing step, behaviour planning integrates diversity directly into the planning process, allowing users to reason about, compare, and select among qualitatively different behaviours. However, this comes with the cost of designing and implemented the diversity dimensions. Regarding future work, one promising direction is the automated or semi-automated construction of behaviour spaces, potentially by learning informative features from data or by incorporating user feedback to iteratively refine dimensions.

\newpage
\bibliography{main}


\end{document}